\documentclass[conference]{IEEEtran}
\IEEEoverridecommandlockouts
\usepackage{cite}
\usepackage{amsmath,amssymb,amsfonts}
\usepackage{algorithm}
\usepackage{algpseudocode}
\usepackage{graphicx}
\usepackage{textcomp}
\usepackage{url}
\usepackage{xcolor}
\def\BibTeX{{\rm B\kern-.05em{\sc i\kern-.025em b}\kern-.08em
    T\kern-.1667em\lower.7ex\hbox{E}\kern-.125emX}}
\begin{document}

\title{Achieving Connectivity Between Wide Areas Through Self-Organising Robot Swarms Using Embodied Evolution}

\author{\IEEEauthorblockN{Erik Aaron Hansen}
\IEEEauthorblockA{\textit{Department of Computer Science} \\
\textit{Oslo Metropolitan University}\\
Oslo, Norway \\
erihanse@outlook.com}
\and
\IEEEauthorblockN{Stefano Nichele}
\IEEEauthorblockA{\textit{Department of Computer Science} \\
\textit{Oslo Metropolitan University}\\
Oslo, Norway \\
stefano.nichele@oslomet.no}
\and
\IEEEauthorblockN{Anis Yazidi}
\IEEEauthorblockA{\textit{Department of Computer Science} \\
\textit{Oslo Metropolitan University}\\
Oslo, Norway \\
anis.yazidi@oslomet.no}
\and
\IEEEauthorblockN{Harek Haugerud}
\IEEEauthorblockA{\textit{Department of Computer Science} \\
\textit{Oslo Metropolitan University}\\
Oslo, Norway \\
harek.haugerud@oslomet.no}
\and
\IEEEauthorblockN{Asieh Abolpour Mofrad}
\IEEEauthorblockA{\textit{Department of Computer Science} \\
\textit{Oslo Metropolitan University}\\
Oslo, Norway \\
asimof@oslomet.no}
\and
\IEEEauthorblockN{Alex Alcocer}
\IEEEauthorblockA{\textit{Deptartment of Mechanical, Electronics} \\
\textit{and Chemical Engineering} \\
\textit{Oslo Metropolitan University}\\
Oslo, Norway \\
alex.alcocer@oslomet.no}
}

\maketitle

\begin{abstract}
Abruptions to the communication infrastructure happens occasionally,
where manual dedicated personnel will go out to fix the interruptions,
restoring communication abilities. However, sometimes this can be
dangerous to the personnel carrying out the task, which can be the case in
war situations, environmental disasters like earthquakes or toxic spills or
in the occurrence of fire. Therefore, human casualties can be minimised
if autonomous robots are deployed that can achieve the same outcome:
to establish a communication link between two previously distant but
connected sites. In this paper we investigate the deployment of mobile ad hoc robots
which relay traffic between them. In order to get the robots to locate themselves appropriately,
we take inspiration from self-organisation and emergence in artificial life,
where a common overall goal may be achieved if the correct local rules on
the agents in system are invoked. We integrate the aspect of connectivity between two sites into the multirobot
simulation platform known as JBotEvolver. The robot swarm is composed of Thymio II robots. In addition, we compare
three heuristics, of which one uses neuroevolution (evolution of neural networks) to show how self-organisation
and embodied evolution can be used within the integration.
Our use of embodiment in robotic controllers shows promising results and
provide solid knowledge and guidelines for further investigations.
\end{abstract}


\section{Introduction}
The advent of Internet of Things (IoT) allows more and more
objects around us being enhanced for communicating
their state and the state of their environment, as well as providing interfaces
for how humans to manipulate them. 
No matter the IoT devices’ purpose, what is common for them is
that they require an underlying backbone connection in order to fully
serve their purpose. Without an established link there is no way of
communicating with the intended users, whether it is for receiving
instructions or giving sensory feedback. This might be because of the way
the system is designed, where the devices are not supposed to be connected
at all times, or if there is a failure somewhere in the network infrastructure
intended for the device. In any case, the link must somehow be established.
Although this could be done manually by humans, in some cases there exist
motivation for doing this in a more autonomous fashion. This could be the
case where the environment has toxic characteristics, or as an aid for firstresponders
after an urban disaster as part of a search and rescue mission.
Ad-hoc networks show promising results for solving some of these
issues, but this requires the agents to be mobile to cover unconnected areas.
Swarm robotics can help us create a backbone for communication exchange
where the robots act as intermediate relay nodes for large area coverage.
By using self-organising robots we can achieve an autonomous solution
to providing network availability in dangerous environments, averting
humans coming to harm.

How can we autonomously create a self-organising network of robots to obtain connectivity
between unconnected end-points?

\begin{figure}
\centering
\includegraphics[width=0.5\columnwidth]{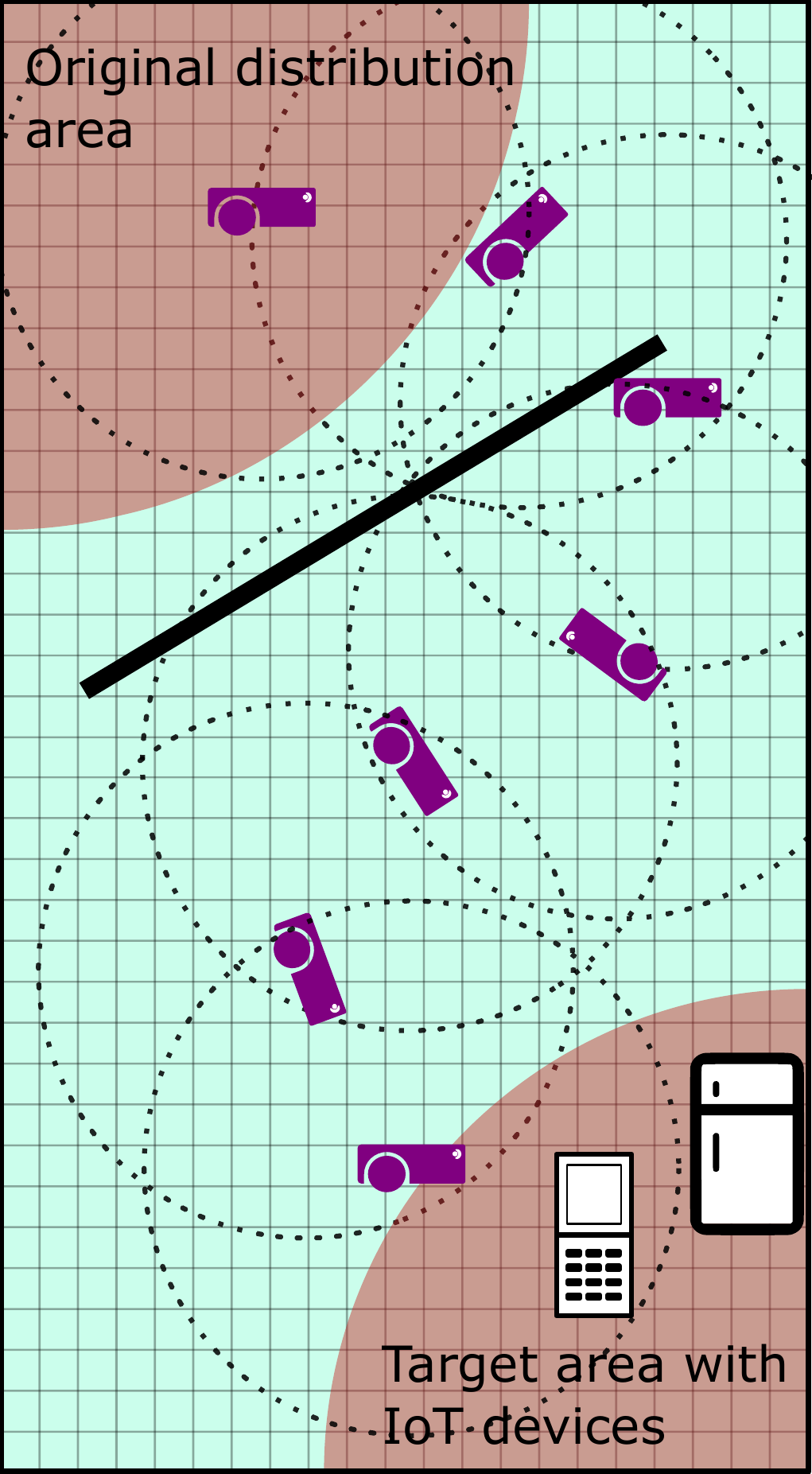}
\caption{Illustration of overall scenario}
\label{fig:intro-problemillustration}
\end{figure}

A self-organising network of robots is a network of mobile agents where
the agents are capable of interacting with each other and their environment.
The robotic agents are equipped with sensory capabilities and can
use actuators to make decisions. The actuators compromises any action
the robots can take, e.g. 
steering and moving in the environment if they have wheels or other forms of
mobile actuators. With a network of robot we achieve a swarm of robots
which collectively can carry out missions better than what any single robot
can achieve, by collaboration.
Such a mission can be establishing a backbone connection with a
distant area by creating a mobile ad-hoc network (MANET), so that we can
communicate and read sensory data with potential IoT devices at the target
area, as illustrated in figure~\ref{fig:intro-problemillustration}. In a MANET there is no preexisting
network infrastructure, meaning there is no central access point
each node in the network must connect to in order to establish connectivity.
This makes the network a decentralised wireless network, where each node
is participating in transmitting packets, as a relay network node. If we
imagine each node being a robot who is aware of one another and of
the environment around it, by dispersing the robots throughout the
environment we should be able to establish a network connectivity from
the original point of dispersion to a target area if the robots collaborate.
This will in turn form a multi-hop routing network connecting
the original deployment area with target area.
In this work, the robots will perform embodied
evolution to adjust their controllers in order to carry out the required task. 

According to \cite{bredeche_embodied_2017}, embodied evolution is a paradigm where
we have the concept of evolution in multi-robotic systems that are:
\begin{itemize}
	\item Decentralised: There is no leader coordinating the evolutionary process of generating and selecting offspring. It is up to the robots themselves to execute this locally, therefore the evolution is \emph{embodied} within the robots themselves.
	\item On-line: Evolution happens while deployed and not off-line, meaning they are not preprogrammed to carry out the task.
	\item Parallel: We have a population of robots performing actions, evolving at the same time and place.
	\item Mating: An action where two or more robots decide to send and/or receive genetic material.
\end{itemize}

The paper is laid out as follows: background information is provided in Section II. Section III describes the used methodology and Section IV outlines the experimental setup. Results and discussion are provided in Section V. Section VI concludes the work and provides direction for future works.

\section{Background}

One of the many challenges within swarm robotics is maintaining versatility regarding the different environments the robots will function in. Robots can be trained to do a task in a particular environment well, but if changes in the surroundings are introduced, for instance if new obstacles appear or the swarm is deployed into another setting, the swarm might not be able to generalise in a sufficient manner. In addition, if the robot experiences some sort of failure, for instance motor fault, it will in many cases be unable to compensate. This motivates the use of embodied evolution, where controllers can be updated online while the robot is deployed.

As to our knowledge the first use of embodied evolution in robotics can be found in~\cite{watson_embodied_1999}, where they trained a robot controller to navigate to a lamp emitting light centered in the middle of an arena, a task commonly known as \emph{phototaxis}. The controller was a simple neural network consisting of one input, a predicate saying which of the two proximity sensors located on the robot had the highest value.

In nature, it appears that ants have carried out connectivity tasks such as congestion control ~\cite{prabhakar_regulation_2012}, ~\cite{brabhakaranternet}. In addition, ants are not the only species studied for their collective behaviour. Slime mold performing network routing is described in~\cite{li_slime_2010}, plants performing collective decision making in~\cite{falik_rumor_2011} and cells doing cycle transitions to solve approximating majority in~\cite{cardelli_cell_2012}. A more comprehensive list can be found in~\cite{navlakha_distributed_2014}.

A classification of evolving robot controllers is given in the next section.

\subsection{Classification of Evolving Robot Controllers}

Online evolution of robotic controllers can be classified into three different categories, according to \emph{when}, \emph{where} and \emph{how} it happens~\cite{eiben_embodied_2010}:
\begin{itemize}
	\item when: off-line vs. on-line, while deployed
	\item where: on-board vs. off-board
	\item how: encapsulated or centralised manner vs. distributed manner
\end{itemize}
In off-line evolution the controllers are developed before the robot is deployed, meaning that the evolutionary operators are only applied before deployment. On-line evolution is the opposite, here controllers are adjusting while deployed. It is also possible to do both, where for instance controllers can be pre-evolved within simulation before deployment in real hardware. Recall however that on-line evolution is a prerequisite for the method to be ``embodied''.

\begin{figure}
\centering
\includegraphics[width=0.6\columnwidth]{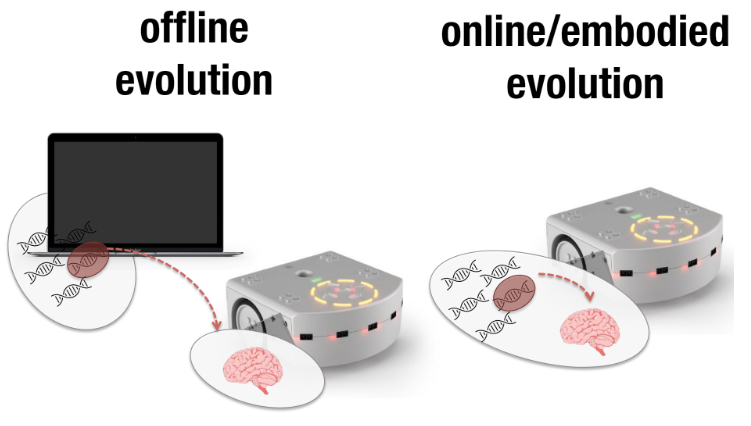}
\caption{The evolutionary algorithm can be executed offline or online. From~\cite{heinerman_ci-group/neatthymio_????}}
\label{fig:emb-offline-online}
\end{figure}

Evolution can take place both off-board and on-board, as illustrated in Figure~\ref{fig:emb-offline-online}. In an off-board situation, an external computer can process fitness-data sent from on-line robots. In this case the execution of the evolutionary algorithm like selection, mutation and recombination is not performed by the robot themselves, but fitness data obtained within the robots is sent to the external machine. After an evolutionary cycle is complete, the external computer injects the robots with new, updated controllers.

\begin{figure}
\centering
\includegraphics[width=0.6\columnwidth]{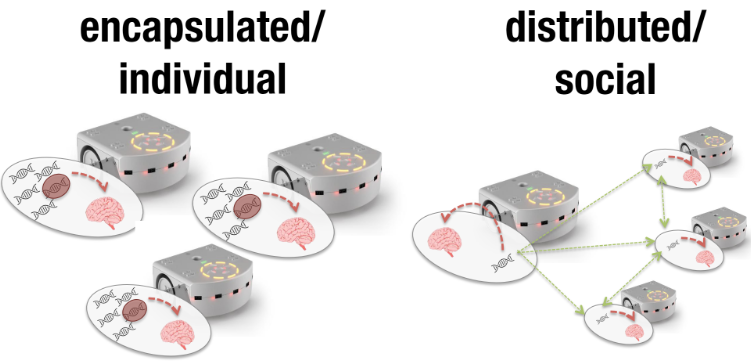}
\caption{Encapsulated vs. distributed maintaining of genomes. From~\cite{heinerman_ci-group/neatthymio_????}}
\label{fig:emb-encapsulated-distributed}
\end{figure}

There are mainly three ways how evolutionary operators are managed: \emph{encapsulated}, \emph{distributed} and a hybrid approach between the two, illustrated in Figure~\ref{fig:emb-encapsulated-distributed}. In the \textbf{encapsulated} version, each robot maintains an internal population of genomes. Certainly, at any given moment only one genome is selected to operate as the controller. The evolutionary process involves the gene pool that resides within each robot, \emph{separately}. There is in other words no interactions with other robots and therefore no genome exchange. In the \textbf{distributed} version, we have local interactions among the robots, meaning they can learn from each other. However, there is only one single genome in every robot's gene pool, so if the environment is big and there are few interactions, the robots will suffer since no optimisations are done. Therefore, a \textbf{hybrid} approach can involve both a bigger internal gene pool like found in the encapsulated version, in addition to exchanging genomes with interacting robots. Also here, if we recall back to the definition of embodied evolution, we need to have the prospect of \emph{mating} in order for the method to be embodied, so a distributed or hybrid approach is required in that sense.

\subsection{NeuroEvolution of Augmenting Topologies}

In 2002 Stanley and Miikkulainen proposed a methodology named NeuroEvolution of Augmenting Topologies (\emph{NEAT})~\cite{stanley_evolving_2002}. Here each connection between perceptrons are explicitly represented in the encoding scheme (the genotype). Initially the neural network starts out like a regular ANN, but later new perceptrons may spawn and new connections between perceptrons may arise, and some be taken away. With time the topology might increase in complexity suiting the task at hand. Another interesting thing about NEAT is that since the network is constantly evolving, new tasks can be learnt while retaining learnt behaviour for previous tasks.

\subsection{Online Decentralised NEAT}

There are different variations of NEAT, and for the purpose of multi-robot systems the adapted NEAT methodology Online Distributed NeuroEvolution of Augmenting Topologies (\emph{odNEAT}) is particularly relevant~\cite{silva_odneat:_2014}.\ odNEAT is a distributed and decentralised neuroevolution algorithm that evolves both weights and network topology in autonomous robots. In the original paper odNEAT is used to solve three tasks: aggregation, integrated navigation and obstacle avoidance, and phototaxis, approximating the performance of rtNEAT, a centralised method and outperforming IM-($\mu$ + 1), a decentralised neuroevolution algorithm. The neural network topology in each robot is augmented progressively, starting out from the simplest possible topology where all inputs are connected to all neurons. Each robot manages an internal population of robotic controllers, having only one deployed at a time. With time, the controllers will complexify, and should suit the task at hand.

\begin{figure}[tbp]
	\centering
	\includegraphics[width=0.8\linewidth]{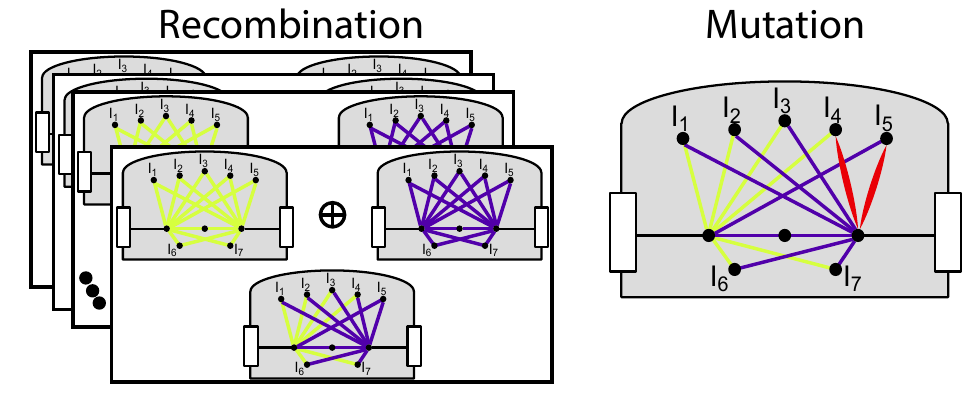}
	\caption{Possible variation operators in embodied evolution}
	\label{fig:emb-thymio-wt-weighs-evoed}
\end{figure}

In order to avoid similar controllers performing poorly, a \emph{tabu list} is used. The tabu list keeps track of recent deployed controllers, so only controllers who are dissimilar in topology will be added.

odNEAT executes locally on each robot, making it a decentralised approach. The general algorithm can be found in Algorithm~\ref{alg:odneat}, and a visualisation in figure~\ref{fig:emb-thymio-wt-weighs-evoed}.

\begin{algorithm}
	\caption{Pseudo-code of odNEAT}
	\label{alg:odneat}
	\begin{algorithmic}
		\State $genome \gets create\_random\_genome()$
		\State $controller \gets assign\_as\_controller(genome)$
		\State $\mathit{energy \gets default\_initial\_energy}$
		\Loop
			\If{$broadcast?$}
			\State $send(genome, robots\_in\_communication\_range)$
			\EndIf
			\If{$has\_received?$}
				\ForAll{$c$ \textbf{in} $received\_genomes$}
					\If{$tabu\_list\_approves()$ \textbf{and} $population\_accepts(c)$}
						\State $add\_to\_population(c)$
						\State $adjust\_population\_size()$
						\State $adjust\_species\_fitness()$
					\EndIf
				\EndFor
			\EndIf
			\State $\mathit{operate\_in\_environment()}$
			\State $\mathit{energy \gets update\_energy\_level()}$
			\If{$\mathit{energy \leq min\_energy\_threshold}$ \textbf{and} $\mathit{not(in\_maturation\_period?)}$}
				\State $add\_to\_tabu\_list(genome)$
				\State $\mathit{offspring} \gets \mathit{generate\_offspring()}$
				\State $\mathit{update\_population(offspring)}$
				\State $\mathit{genome \gets replace\_genome(offspring)}$
				\State $\mathit{controller \gets assign\_as\_controller(genome)}$
				\State $\mathit{energy \gets default\_initial\_energy}$
			\EndIf
		\EndLoop
	\end{algorithmic}
\end{algorithm}

\subsection{JBotEvolver - A Simulation Platform for Evolutionary Robotics}

\begin{figure}[!t]
	\centering
	\includegraphics[width=\linewidth]{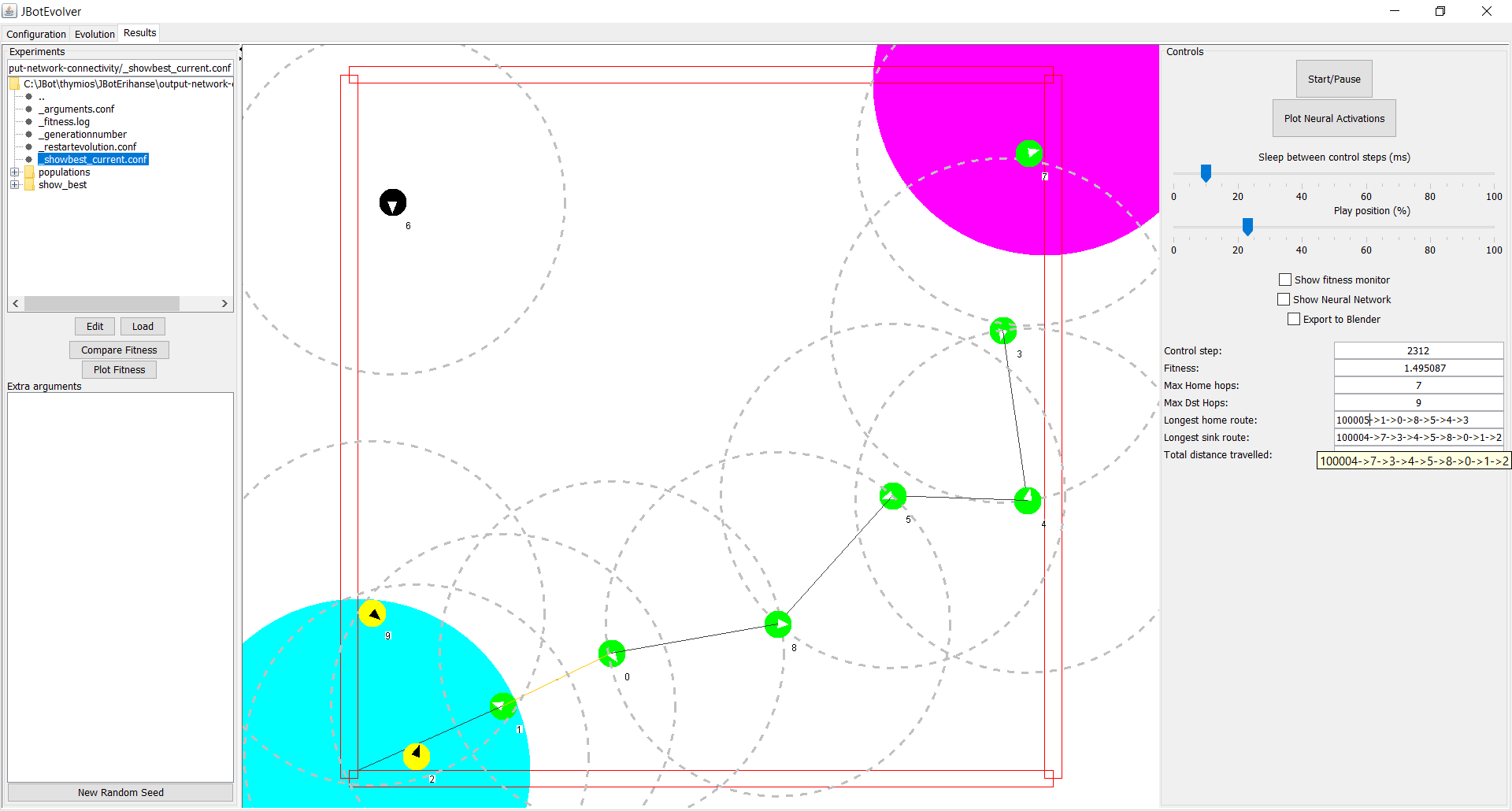}
	\caption{JBotEvolver results view}
	\label{fig:jbot-results}
\end{figure}

JBotEvolver is a versatile simulation platform for evolutionary robotics written in Java~\cite{duarte_jbotevolver:_2014}. The platform is open source and currently maintained under \url{https://github.com/BioMachinesLab/jbotevolver}. The framework also provides a graphical user interface for configuring evolutions and viewing results, as shown in Figure~\ref{fig:jbot-results}. A huge advantage of the simulation platform is that classes are loaded dynamically. This means that if we are debugging the executable through any integrated development environment (IDE) supporting hot code replace like for instance Eclipse or Visual Studio Code, we can start a simulation to observe robots' behaviours, set a breakpoint within the controller and make some changes, and then continue the program execution. We can see the changes immediately without restarting the debug session, providing us with the opportunity of quick and incremental development.

In the configuration tab we can set numerous parameters to use. Not all parameters are compatible with each other. This requires both domain and implementation knowledge of the code base. The main parameters are:
\begin{itemize}
	\item Robot: Here we set parameters for the robots, like the colour, size between wheels, number of robots and so on.
	\item Sensor: We can add additional sensors.
	\item Actuator: We can add actuators to the robots, for instance wheels, LED lights, speakers.
	\item Controller: Set the controller controlling the robot.
	\item NeuralNetwork: Set if the robot should use a neural network as the controller. Possible implementations use NEAT, Continuous time recurrent neural networks (CTRNN) and Multilayer Perceptron Networks.
	\item Population: Related to managing evolved controllers.
	\item TaskExecutor: Whether to run simulations in parallel, synchronously or by using a distributed platform called Conillon.
	\item Evolution: Which strategy to use to evolve individuals in the population.
	\item EvaluationFunction: Which evaluation function to use to determine the performance of the robots deployed.
\end{itemize}

\section{Methodology}

\begin{figure}[!t]
	\centering
	\includegraphics[width=\linewidth]{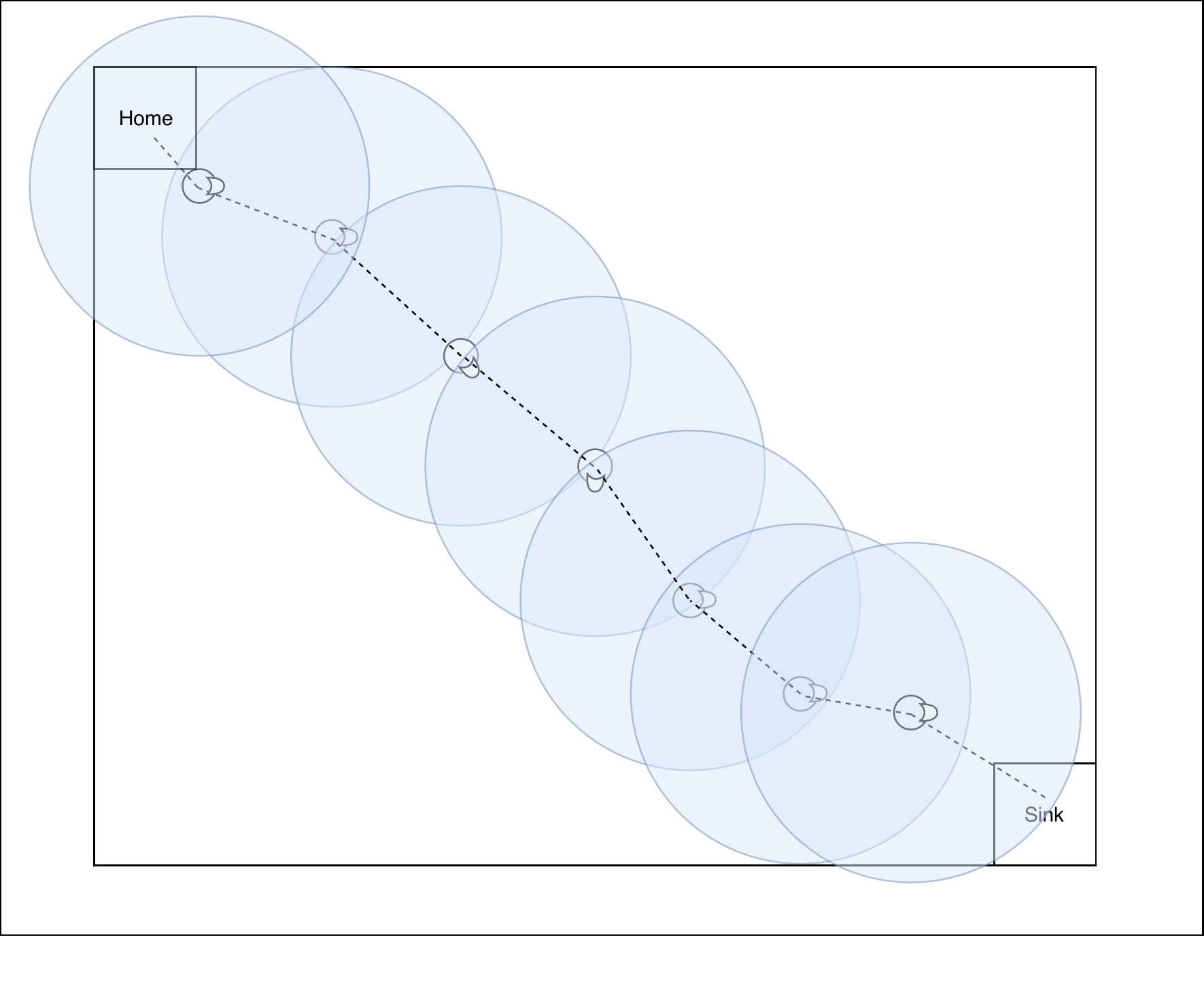}
	\caption{Illustration of components involved in task being solved.}
	\label{fig:fig-met-task}
\end{figure}

The goal of the work herein is to connect two distant access points using an ad-hoc network of mobile robots, creating a relay network for communication between the two points. An illustration of the task can be found in Figure~\ref{fig:fig-met-task}. In the figure we can see different kind of geometric shapes. In the corners there are two boxes which represent stationary connection points, denoted as \emph{Home} and \emph{Sink}. Near to Home is where the robots start initially, and it is up to them, or more specifically, their controllers to make them self-organise so that they move into place, together relaying information between the two distant points. Furthermore, each robot is equipped with a networking interface so that it can transmit packets containing information with other network devices in range of the robot. This gives us a dynamic system with moving and communicating agents. We also see that the robots are within a square, which represents walls, meaning the robots are limited regarding where they can move. With the walls in place, this sort of environment can be referred to as an arena environment, with a given width and height.

To determine whether we can accomplish our goal, we will present three methods depending on what controllers the robots are equipped with:
\begin{itemize}
	\item method using random-walk controllers
	\item pre-programmed controller
	\item variant using embodied evolution as the controller
\end{itemize}

We need the robots to coordinate themselves among each other to connect the two access points of interest. Since they relay messages through wireless communication, two robots need to be within \emph{signal range} with one another. Therefore, in the real world we would maybe be more interested in the \emph{signal strength} between the robots than the actual range. In reality, a direct network connection between two agents is continuous, meaning that the further away a distant robot is located from a given robot, the weaker the signal strength between the two. We are more interested in the self-organisation aspect than network technicalities, so which protocols can be used in an ad-hoc network, how routing would be done in said network, and so on is certainly of interest, but out of scope for this paper. Thus the main focus will be on the placements of the individual robots according to each other more than how they communicate.

Each robot does not know anything else than what it can sense from the outside world, although each robot is allowed to relay communication traffic from connected robots, and use this in their decision making. One information a robot knows from the outside world is which robots it is adjacent with, i.e. within communication range. 

In addition, each robot has a \emph{route} to both home and sink. Every robot has their own route to home and sink, and we will refer to these as the \emph{home route} and \emph{sink route}. These routes are linear lists, with information about the adjacencies of the contained robots. This means given a robot's route \(R_h = \{home, robot1, robot2, robot3\} \), robot1 is connected to robot2, and robot2 is connected to robot3. In other words, an element is adjacent with the one before and after it. The route is learnt through adjacent robots. It compares its own route home, and if a neighbour has a shorter route than itself, it updates its route home, so that there is a shortest path first (SPF) connection with home.

Related to connectivity, a robot knows the following:
\begin{itemize}
	\item which neighbours it has
	\item the route home
	\item the route to the sink.
\end{itemize}

\subsection{Random Walk Controller}
The first of the three controllers we test on the robots is the random walk controller. Its behaviour is quite intuitive: Every $x'th$ time~step, pick a random direction between left, right, forward and backward, and apply output to the wheels accordingly. Algorithm~\ref{alg:random-walk-controller} shows the basics in the algorithm. By driving left or right in this case does not necessarily mean: apply appropriate speed on the two wheels so that the robot rotates 90 degrees from current standpoint. It just means: run for a given time leftwards or rightwards. This might end up making the robot turn partially to any direction. The main idea is simply to make the robot drive around randomly. Given the random behaviour of the robots and the fact that the environment is within a closed area, we will get a somewhat evenly distribution of the robots throughout the environment over time. If the robots have long enough range, the area is small enough, or we have enough robots deployed, we should achieve connectivity between Home and Sink simply by chance. We will use this as a reference to compare with the other two methods.

\begin{algorithm}[H]
	\caption{Random Walk Controller}
	\label{alg:random-walk-controller}
	\begin{algorithmic}
		\Function{$controlstep$}{$time$}
		\If {$\mathit{full\_connection?}$}
		\State $stop()$ 
		\Else
		\State $random\_walk()$
		\EndIf
		\EndFunction
		\Statex
		\Function{$\mathit{random\_walk()}$}{}
		\State $\mathit{direction \gets random(left, right, forward, backward)}$
		\State $drive(direction)$
		\EndFunction
	\end{algorithmic}
\end{algorithm}

\subsection{Pre-programmed Approach (``longest home route'')} 
The second approach is slightly more advance than the previous. The general idea is to divide the overall goal of connecting home and sink into smaller sub-tasks, by building up a chain stretching out from home. As robots connect to the chain, they become stationary and a part of the chain themselves so that future robots can continue the chain. Eventually as the chain is formed we might reach the sink, making home and sink a connected component via the chain. If the chain is somehow broken, robots disconnected to the longest home route will continue looking for the route.

The chain itself is defined through the notion of the \emph{longest home route}. This is the overall longest routes of SPF routes stretching out from home. Since this is in terms of hops, it is also probable that this route contains at least one robot that is further away from home than other robots connected to home. A robot will add itself to this chain, as it would do with any chain, if this is the shortest home route for the robot. The special case now, is that the longest home route now has a new node within its list of nodes, resulting in a chain extension. There is one additional rule to fulfill to be part of the longest home route: that all neighbouring nodes that is also part of the longest home route must be within a predetermined interval. 

\begin{algorithm}[H]
	\caption{Preprogrammed controller}
	\label{alg:preprogrammed}
	\begin{algorithmic}
		\Function{$controlstep$}{$time$}
		\If {$\mathit{full\_connection?}$}
		\State $stop()$ 
		\ElsIf {$part\_of\_longest\_home\_route?$ and $in\_optimal\_range?$}
		\State $stop()$
		\Else
		\State $random\_walk()$
		\EndIf
		\EndFunction
	\end{algorithmic}
\end{algorithm}

\subsection{Evolvable Approach Using odNEAT}
Lastly we equip the previous controllers with odNEAT. 

Previously we have had robots running randomly around in the environment, and we have had the pre-programmed controllers with the additional ability to build up a chain. By experimenting through visual observations of the two different methods, it is apparent that the robots seem to crash often, both with the surrounding walls and with each other. Additionally, the robots do not need to be in direct contact with each other, since they are doing wireless transfer, i.e. having a spatial communication range. As seen in Figure~\ref{fig:fig-met-task}, if robots investigate the bottom left and top right corner, they are covering uninteresting areas. They also need to be spread out in order for the chain to reach the sink, so they need to have a certain distance between each other. Another aspect to mention is that sensory inputs can be given to the controller, so the controller can determine what power to put on the wheels based on what the robot can sense from the environment.

Thus, given the observed behaviour when using the two simpler controllers, and our intuitive understanding of how the solution should form itself, we equip the robots with odNEAT algorithm to alleviate the described problems. This will be done by penalising controllers leading to a robot crashing, and punishing controllers not seeming to be part of the longest home route. The notion of ``longest home route'' is still used, doing a full stop when partaking in it, but trying to avoid the negative traits mentioned above.

\section{Experimental Setup}

We use the extension of the JBotEvolver robotic simulation framework as toolset, and test our methods in simulation. For each method, we run 30 simulations each with 10, 15 and 20 robots, with two different environment dimensions as seen in Table~\ref{tab-experiments}. The dimensions are in simulation meters, so 2x5 implicates an environment with a 2 meters width and 5 meters height. The broadcast range is only viable for the odNEAT controller, and specifies at what ranges robots are able to communicate with each other for transmitting genomes. Note: It does not affect the home route calculation, that is still using the \emph{intended} communication range, as indicated by the dotted circle around each robot. The reason for the communication range being infinite is the implementation of the odNEAT extension in JBotEvolver. In reality, datagram sockets are used and the robots communication range would be limited by their built-in network interface hardware capabilities. 

All of the experiments are run on a laptop running Windows 10, and compiled using Java™ SE Development Kit 8, Update 144 (JDK 8u144). The source code for the experimental framework and experimental simulations is available open sourge on GitHub \footnote{\url{https://github.com/erihanse/JBotEvo/}}, and was developed as part of the first author's master thesis project.

\subsection{The Robot}
\label{sec:exp-robot}

Each robot knows about its environment around it. Since this is a simulation, some simplifications have been taken in the way that a robot gets to know who its neighbouring robots are, by making a call to the environment. The simulated robots are Thymio II, which include seven InfraRed (IR) proximity sensors (five arranged in the front and two on the back of the robot) and two wheels.

\subsection{The Environment}
\label{sec:exp-environment}

\begin{figure}[!t]
	\centering
	\includegraphics[width=0.6\linewidth]{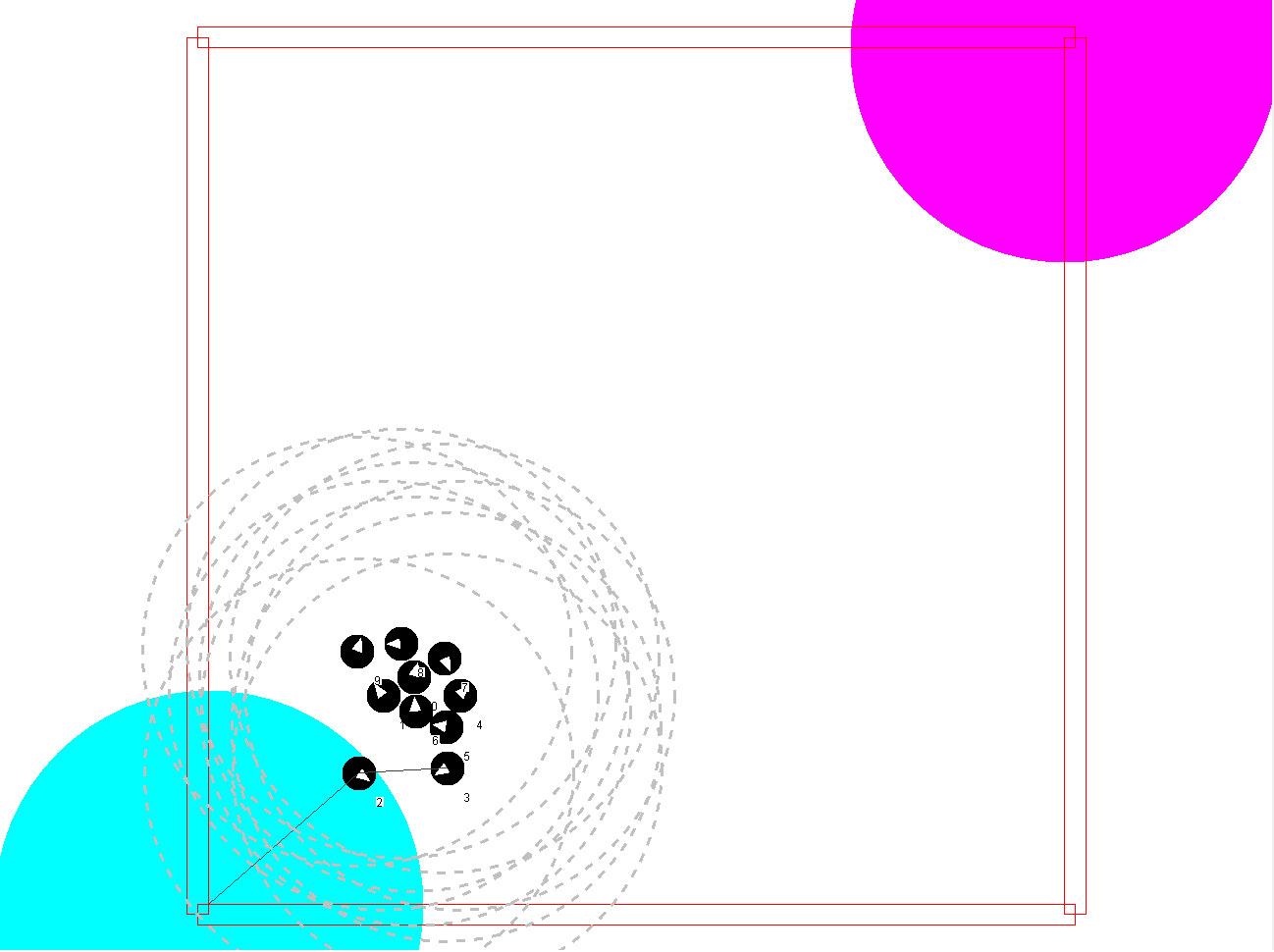}
	\caption{Initial placement in a 4x4 size environment.}
	\label{fig:met-4x4-env}
\end{figure}

\begin{figure}[!t]
	\centering
	\includegraphics[width=0.6\linewidth]{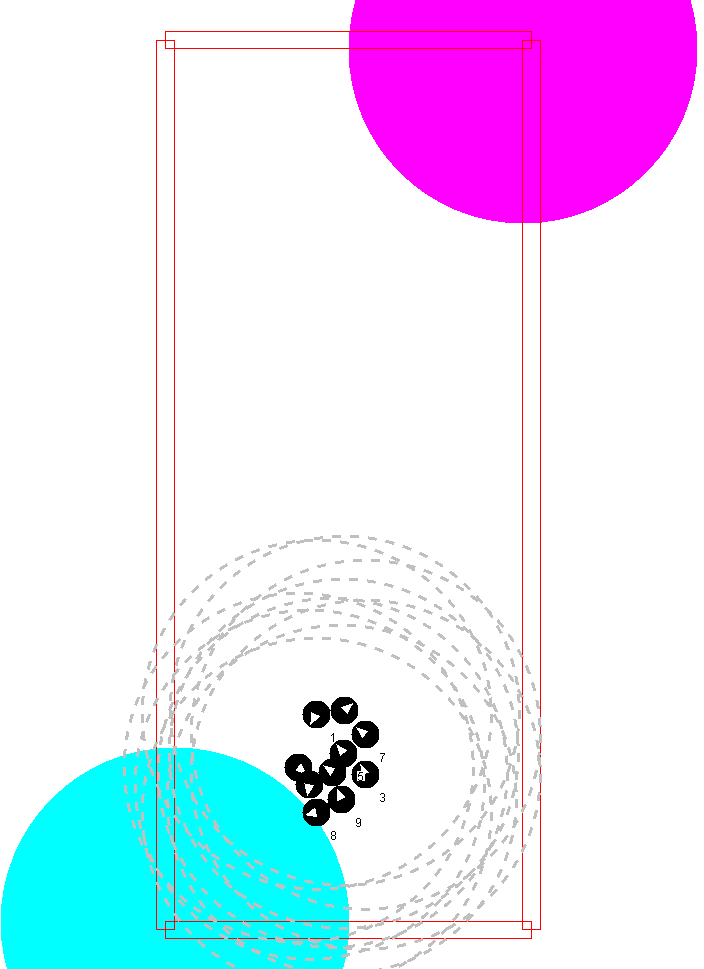}
	\caption{Initial placement in a 2x5 size environment.}
	\label{fig:met-2x5-env}
\end{figure}

The environments and initial robot placement for the two different environments can be seen in Figure~\ref{fig:met-4x4-env} and~\ref{fig:met-2x5-env}. Every simulation can run up to maximum 10'000 simulation steps, where 10 steps corresponds to one second in simulation. This implicates that if a set of robots fail to find a solution within the given time span, the simulation will be cut off, meaning that if time runs out it will be reported that a solution was found at 10'000th time step.

Since the concept of \emph{the longest home route} is central in our approaches, we will add the capability of seeing the IDs of the robots partaking in the route, as well as rendering a line between these robots.

\begin{table}[h]
\centering
\begin{tabular}{|l|l|}
\hline
Controllers used       & Random Walk Controller, Home Route \\ 
					   & Controller and odNEAT Controller \\ \hline
Group size             & 10, 15 and 20 robots    \\    \hline                           
Environment size       & 2x5 meters and 4x4 meters \\ \hline                  
Simulation duration    & 10'000 steps (1000 seconds) in simulation \\ \hline
Runs per configuration & 30 \\  \hline
Broadcast range        & $\infty$ \\ \hline 
Maximum speed          & 0.20 m/s  \\  \hline                                           
\end{tabular}
\caption{Overview of experiments. Configurations for 3 different controllers, 3 group sizes and 2 environment dimensions yields in total 18 experiments. Each experiment will be run 30 times.}
\label{tab-experiments}
\end{table}

\subsection{Visualisation of Neural Network Topologies}
\label{sec:met-visualisation}
In addition to what has been mention above, we provide a visualization of the evolved neural networks, to ensure that they do evolve according to the basic principles of NEAT. It is expected that initially all input neurons are connected to every output neuron. JBotEvolver provides a tick box so that if checked, a window will display input and output neurons, their connections and the activations functions. For this purpose the Graph visualisation software \emph{Graphviz}\footnote{\url{http://www.graphviz.org/}} has been used. However, the existing implementation did not include multi-layer topologies, so their implementation is adjusted to also support arbitrary topologies. We therefore launch a simple simulation in the GUI to plot the neural network of a single robot at the start and end of the simulation.

\section{Results and Discussion}

First, results regarding running time until a solution was found and distances traveled in same simulation are presented. We also show examples of evolved neural networks at the beginning of the evolutionary cycle and at the end.

We recorded the running time and distance traveled of robots running the three different controllers for robot group sizes of 10, 15 and 20, performing in a 4x4 arena and 2x5 arena.

\subsection{Environment with dimensionality 4x4}
\begin{figure}[!t]
	\centering
	\includegraphics[width=\linewidth]{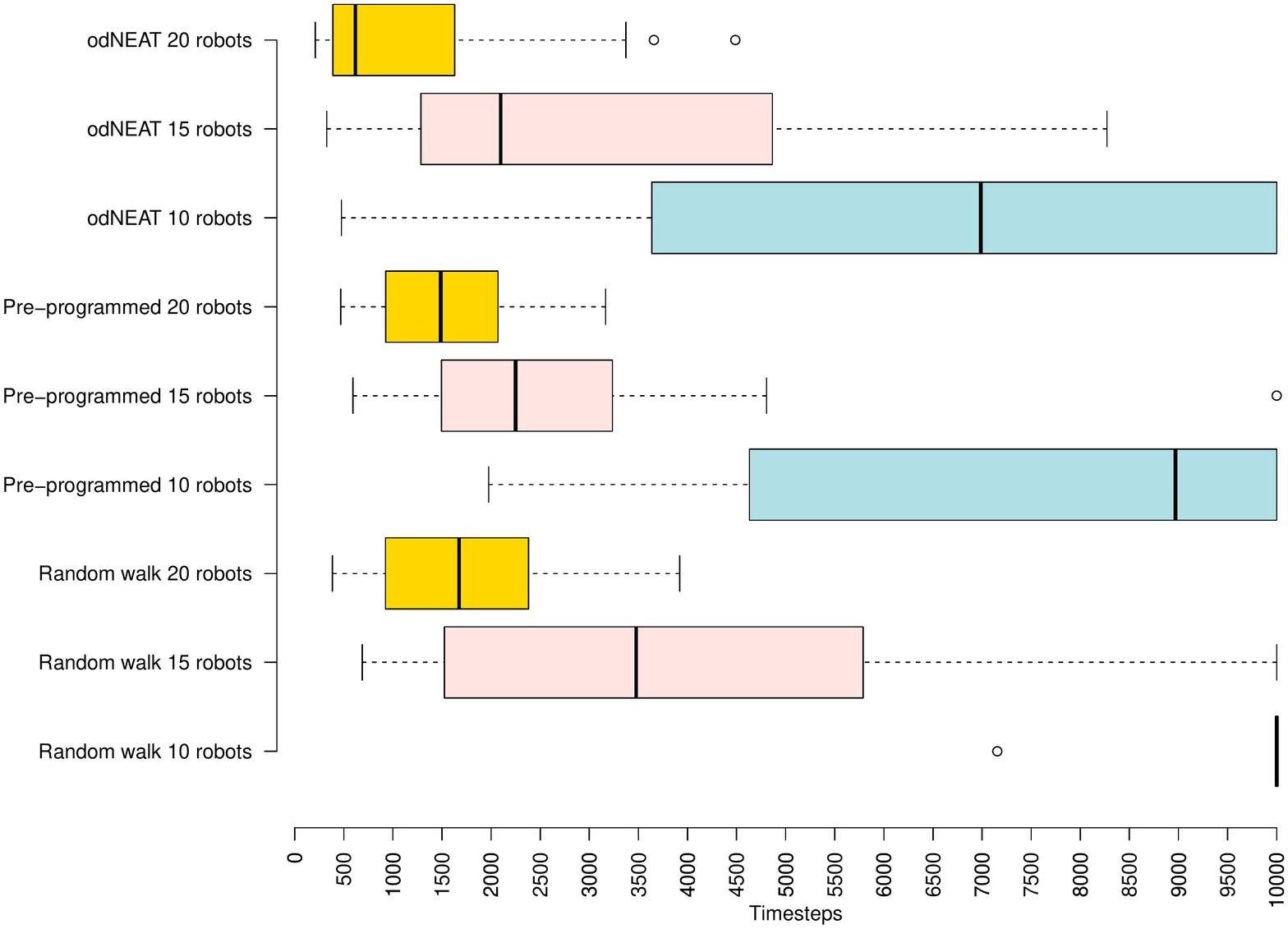}
	\caption{Time taken before full connectivity, environment 4x4}
	\label{fig:res-timesteps}
\end{figure}

\begin{figure}[!t]
	\centering
	\includegraphics[width=\linewidth]{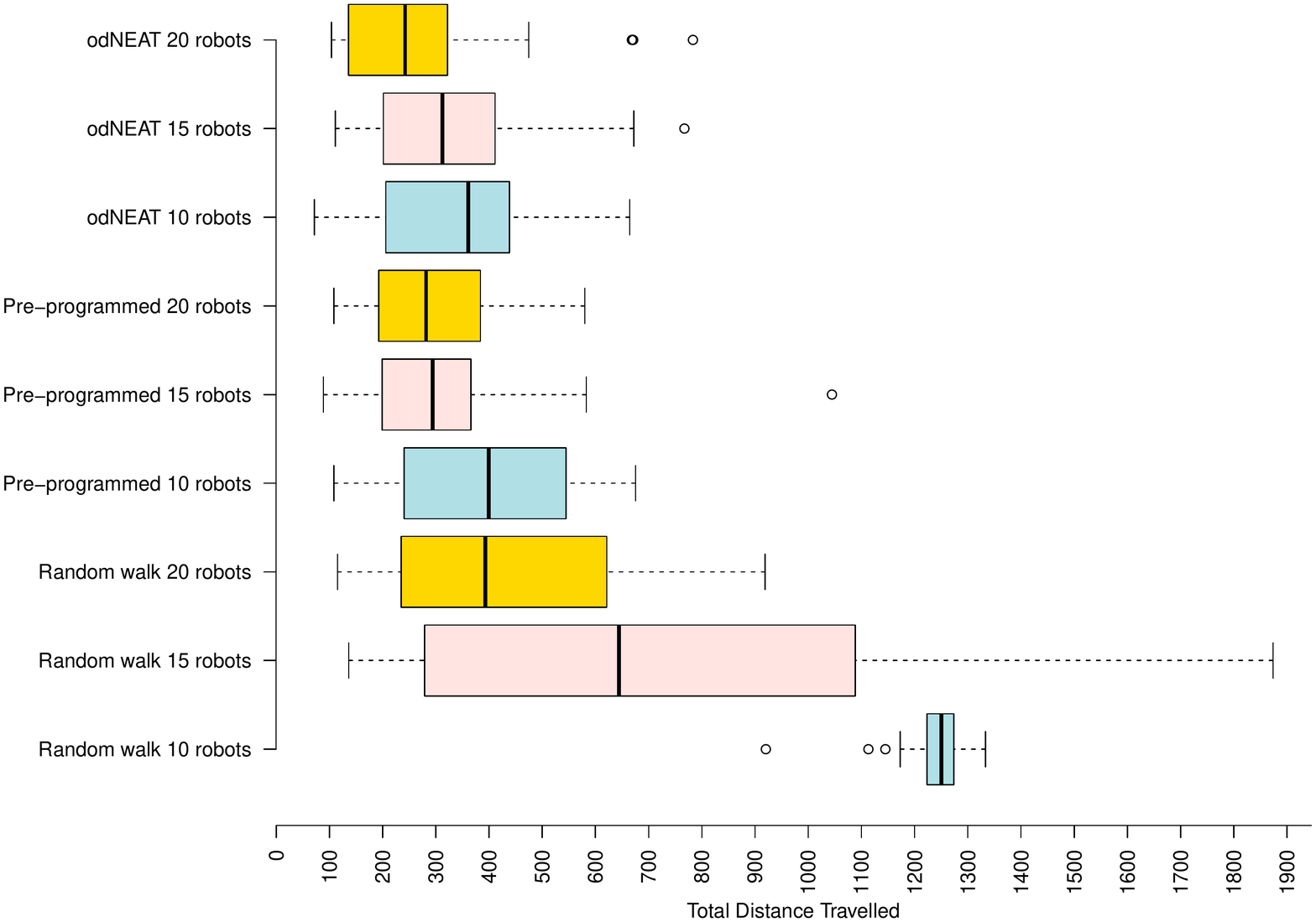}
	\caption{Distances traveled, environment 4x4.}
	\label{fig:res-distance-travelled}
\end{figure}

For the 4x4 environment, results for how many timesteps until solutions were found can be seen in Figure~\ref{fig:res-timesteps}, and the distances traveled in Figure~\ref{fig:res-distance-travelled}.
While odNEAT clearly outperformed the random walk controller, we also compared the odNEAT variant with the pre-programmed approach to produce some statistical measures by doing a welch two sample t-test. The difference between the two methods is statistically significant only for the experiment with 20 robots (p=0.03858), i.e. odNEAT performs better. For 10 and 15 robots, the difference between the two methods is not statistically significant.

\subsection{Environment with dimensionality 2x5}
\begin{figure}[!t]
	\centering
	\includegraphics[width=\linewidth]{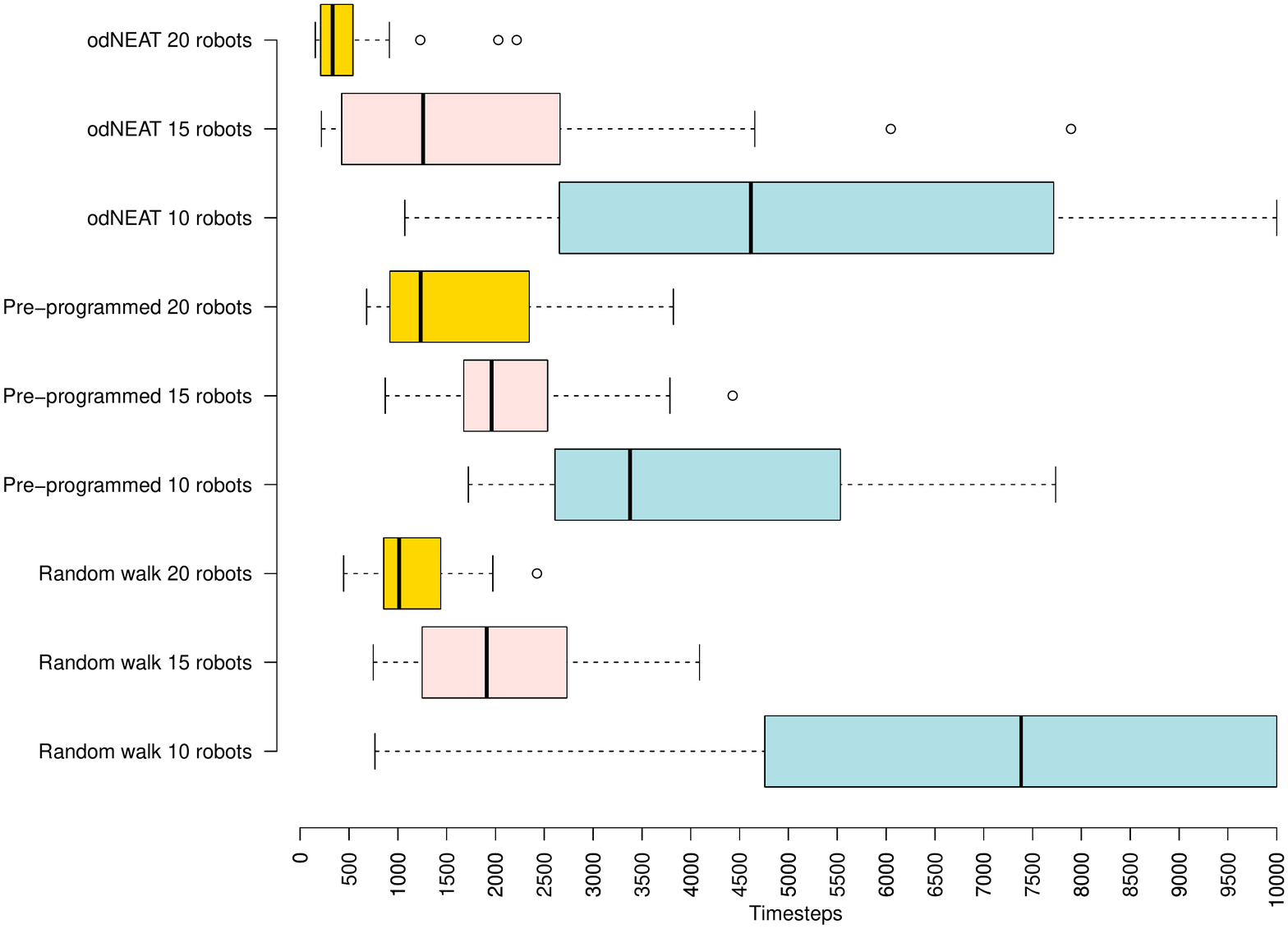}
	\caption{Time taken before full connectivity, environment 2x5.}
	\label{fig:res-timesteps-env2}
\end{figure}

\begin{figure}[!t]
	\centering
	\includegraphics[width=\linewidth]{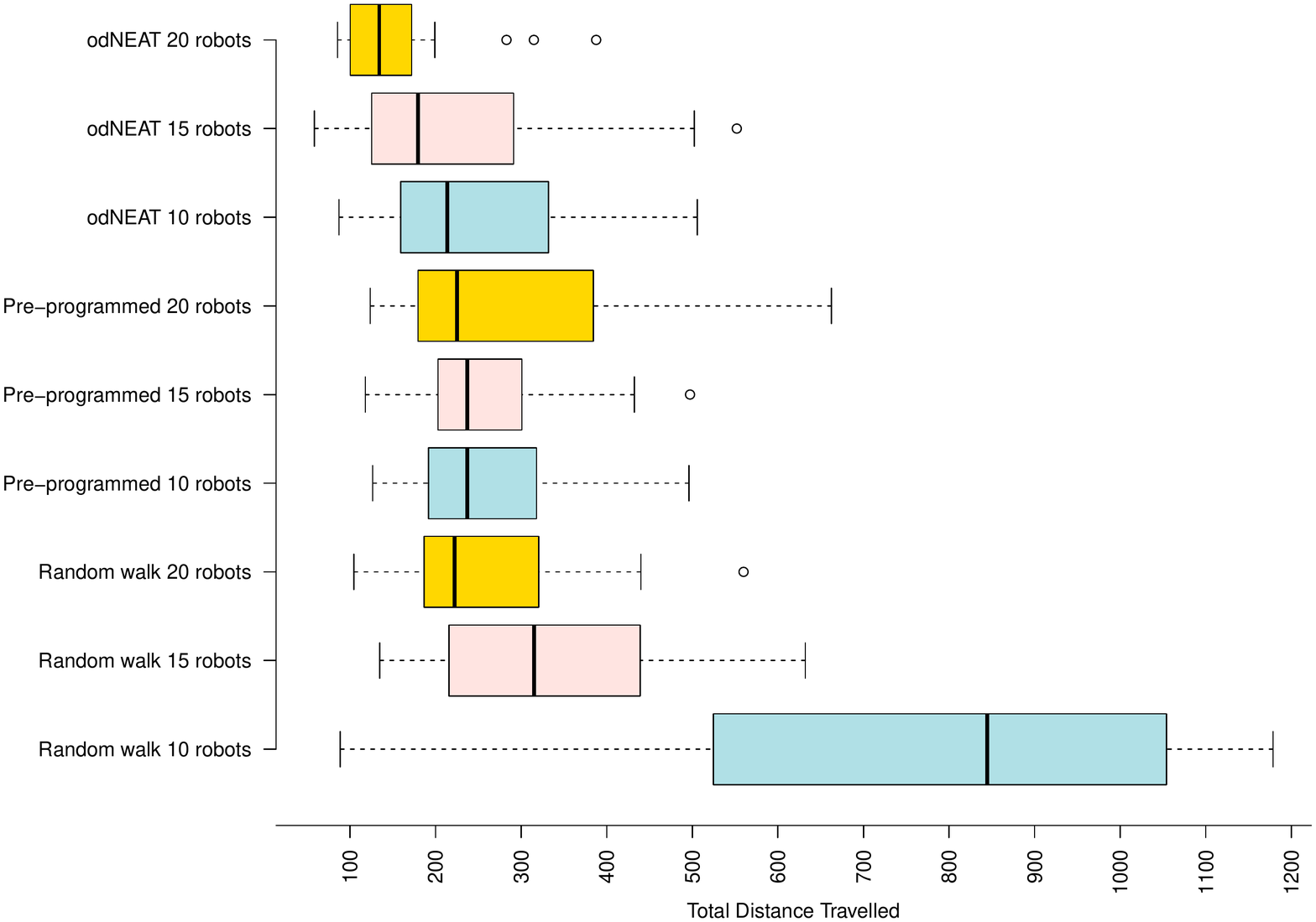}
	\caption{Distances traveled, environment 2x5.}
	\label{fig:res-distance-travelled-env2}
\end{figure}

For the 2x5 environment, results for how many timesteps until connectivity achieved is presented in Figure~\ref{fig:res-timesteps-env2}, and the accumulated distances traveled in Figure~\ref{fig:res-distance-travelled-env2}. We also do a  welch two sample t-test comparison here, as we did in previous section. The difference between the odNEAT variant with the pre-programmed approach is statistically significant only for the experiment with 20 robots (p=2.603e-07), i.e. odNEAT performs better. For 10 and 15 robots again the difference between the two methods is not statistically significant.

\begin{figure}[!t]
	\centering
	\includegraphics[width=\linewidth]{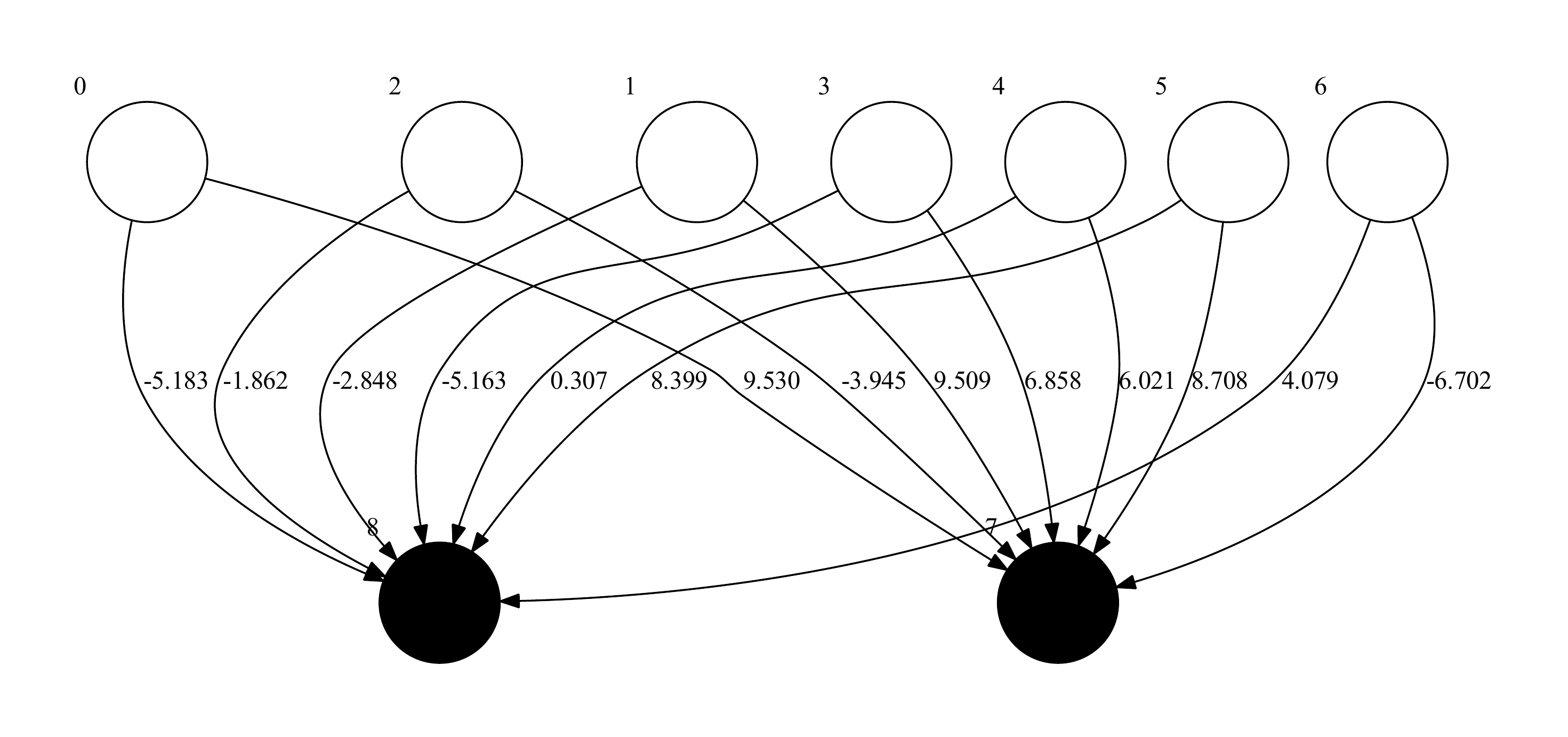}
	\caption{Controller Neural Network Topology of one of the robots at the \emph{beginning} of the simulation.}
	\label{fig:res-graph-start}
\end{figure}

\begin{figure}[!t]
	\centering
	\includegraphics[width=\linewidth]{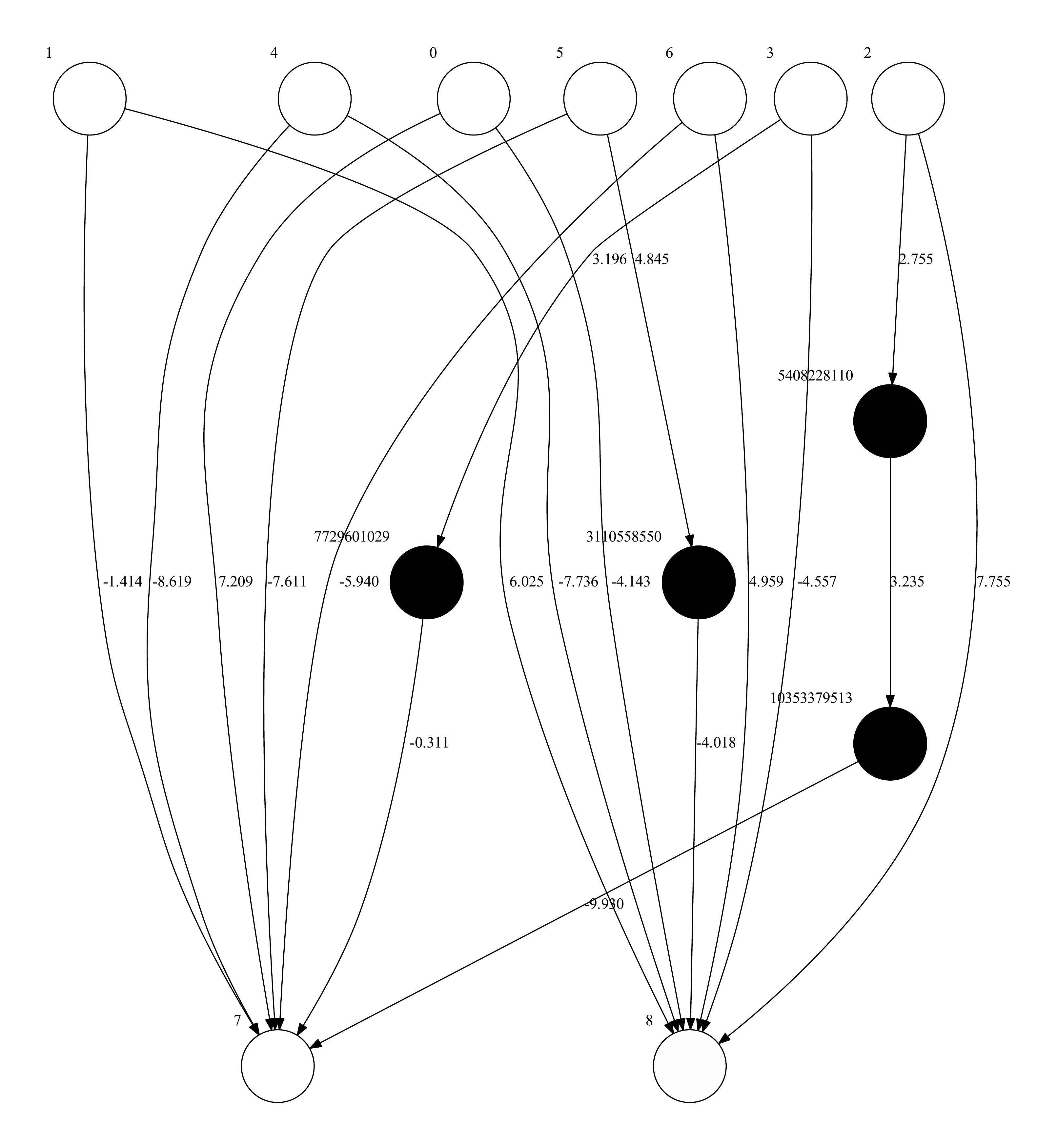}
	\caption{Controller Neural Network Topology of one of the robots at the \emph{end} of the simulation.}
	\label{fig:res-graph-end}
\end{figure}

As example, we observed that in the beginning of the simulation the topology had the form as seen in Figure~\ref{fig:res-graph-start}. At the end of the simulation, the topology had the form as seen in Figure~\ref{fig:res-graph-end}.

\subsection{Discussion}

The pre-programmed approach shewed certainly an improvement from the random walk, so this way of building up a chain by the concept of longest home route was promising. The task at hand is quite difficult, because the agents partaking in the system are supposed to mimic individuals which potentially could be a real self-organising robots. This means the robots cannot know anything else than what they can sense and receive from neighbouring robots. It needs to be mentioned that the placements of the home and sink is done in such a way that no matter what direction the robots moved, it would be more likely that the chain was forming in the correct direction. If the home location was placed in the middle of an environment, the chain could stretch out in the wrong direction.

The first observation that can be done when comparing the results, is that number of robots deployed seem to have a general effect on how long it takes before solutions are found. More robots used gives better performance. This is quite intuitive, but nevertheless worth mentioning. In a square 4x4 environment, the random walk robots never seemed to even find a solution before time ran out when only 10 robots were deployed.

In addition, we observe distances traveled seemingly decreasing as the number of robots is increased, except for with the pre-programmed approach. This also makes sense, because the sooner we find a solution, the sooner the simulation is stopped, and the accumulated distances traveled is registered.

With 10 robots the pre-programmed approach performed better than odNEAT in the 2x5 sized environment, but when increasing the group size to 15 and 20, odNEAT seemingly went ahead. However, due to high p-values from the welch tests, we can only say with statistical significance that our odNEAT based method performed better for the largest group size, independent of the two dimensions tested.

In both of the topologies, we had 7 input neurons and 2 output neurons. The 7 input neurons corresponded with the proximity sensors of the robot, where 1 input neuron was used per sensor. By code inspection we know that input neuron 0-4 were fed values from the 5 proximity sensors in front, and input neuron 5-6 were fed sensory inputs from the proximity sensors on the back of the robot. The two output neurons 7-8 represented the two wheels of the robot. In the beginning, every input neuron was connected to every output neuron. In the end of the evolutionary run, we observed a new topology of the neural network {--} where a deep network with two additional hidden layers is typically achieved.

\section{Conclusion and Future Work}

In this paper, we investigated the use of self-organisation and embodied evolution to achieve connectivity. We used the multi robot simulator JBotEvolver to perform simulations with three different methods involving random walk, a pre-programmed approach with a trivial algorithm, and the use of embodied evolution.
Common for all of the methods applied was the correlation between group size and performance. The more robots applied, the quicker a solution could be found.

An embodied approach gives a promising outline for future exploration of neuroevolution in robotic controllers to achieve network connectivity. Results for smaller group sizes showed a few challenges, however as the number of robots increased odNEAT outperformed random walk and top-down methods.

When running a live simulation (i.e. inspecting one simulation in the GUI and not running several background simulations), we observed that some robots did try to avoid each other, and the walls around. However, some of the robots tend to get stuck in an infinite spinning loop until they ran out of energy. This might be because the robot were equipped with a poorly performing controllers. When only having 10 robots, and assuming 3 of these are running with a poor performing topology, it would require some additional heuristic so that the robots would not waste simulation time, and rather quicker attempt something new. This can be done in several ways {--} more experimental work on what parameters to use for the odNEAT algorithm, different initial energy values, an improved local fitness function, or an attempt to introduce heuristics such as visiting places not previously visited (e.g., novelty search).

Real hardware sensors are certainly not as perfect as what is used in simulations. For instance, proximity sensors may suffer inaccuracies when obstacles are either too far or too close. Therefore the \emph{reality gap} and how to overcome it between sensors used in simulation and reality must also be explored in future works.

\bibliographystyle{ieeetr}
\bibliography{mybib.bib}

\end{document}